%% file: anonymous-submission-latex-2026.tex
\documentclass[letterpaper]{article} 
\usepackage[draft]{aaai2026}
\usepackage{times}  
\usepackage{helvet}  
\usepackage{courier}  
\usepackage[hyphens]{url}  
\usepackage{graphicx} 
\usepackage{natbib}  
\usepackage{caption} 
\usepackage{algorithm}
\usepackage{algorithmic}

\usepackage{newfloat}
\usepackage{listings}
\DeclareCaptionStyle{ruled}{labelfont=normalfont,labelsep=colon,strut=off} 
\floatstyle{ruled}
\newfloat{listing}{tb}{lst}{}
\floatname{listing}{Listing}
\usepackage{times}
\usepackage{latexsym}

\usepackage[T1]{fontenc}
\usepackage[utf8]{inputenc}

\usepackage{microtype}

\usepackage{inconsolata}

\usepackage{graphicx}
\usepackage{makecell}
\usepackage[utf8]{inputenc}  
\usepackage[T1]{fontenc}     
\usepackage{url}            
\usepackage{booktabs}       
\usepackage{amsfonts}       
\usepackage{nicefrac}       
\usepackage{microtype}      
\usepackage{graphicx}
\usepackage{natbib}
\usepackage{times}
\usepackage{soul}
\usepackage{url}
\usepackage{caption}
\usepackage{graphicx}
\usepackage{amsmath}
\usepackage{amsthm}
\usepackage{booktabs}
\usepackage{algorithm}
\usepackage{algorithmic}
\usepackage{longtable}
\usepackage{array}
\usepackage{multirow}
\usepackage{amssymb}
\usepackage{multirow, colortbl, xcolor}
\usepackage{booktabs}
\usepackage{listings}
\usepackage{array} 
\usepackage{subcaption} 
\usepackage{enumitem} 
\newcolumntype{C}[1]{>{\centering\arraybackslash}p{#1}}
\newcolumntype{M}[1]{>{\centering\arraybackslash}m{#1}}
\usepackage{tabularx} 

\usepackage{colortbl} 
\usepackage{xcolor}
\usepackage{siunitx}
\definecolor{headergray}{gray}{0.9} 
\definecolor{groupgray}{gray}{0.95} 
\definecolor{rowaltgray}{gray}{0.98} 

\newcommand{\customsize}[1]{\fontsize{10}{12}\selectfont}
\definecolor{lightgray}{gray}{0.95}
\definecolor{darkgray}{gray}{0.4}
\definecolor{purple}{rgb}{0.58,0,0.82}
\definecolor{blue}{rgb}{0.13,0.13,1}

\usepackage{tocloft} 
\usepackage[utf8]{inputenc} 
\title{Understanding and Optimizing Agentic Workflows \\via Shapley value}

\author{
    \textbf{Yingxuan Yang}\textsuperscript{\rm 1},
    \textbf{Bo Huang}\textsuperscript{\rm 1},
    \textbf{Siyuan Qi}\textsuperscript{\rm 1},
    \textbf{Chao Feng}\textsuperscript{\rm 1},
    \textbf{Haoyi Hu}\textsuperscript{\rm 1},\\
    \textbf{Yuxuan Zhu}\textsuperscript{\rm 2},
    \textbf{Jinbo Hu}\textsuperscript{\rm 1},
    \textbf{Haoran Zhao}\textsuperscript{\rm 1},
    \textbf{Ziyi He}\textsuperscript{\rm 3},
    \textbf{Xiao Liu}\textsuperscript{\rm 4},
    \textbf{Muning Wen}\textsuperscript{\rm 1},\\
    \textbf{Zongyu Wang}\textsuperscript{\rm 4},
    \textbf{Lin Qiu}\textsuperscript{\rm 4},
    \textbf{Xuezhi Cao}\textsuperscript{\rm 4},
    \textbf{Xunliang Cai}\textsuperscript{\rm 4},
    \textbf{Yong Yu}\textsuperscript{\rm 1},
    \textbf{Weinan Zhang}\textsuperscript{\rm 1}
}
\affiliations{
    \textsuperscript{\rm 1} Shanghai Jiao Tong University, China 
    \textsuperscript{\rm 2} University of Chicago, USA \\
    \textsuperscript{\rm 3} University of Toronto, Canada 
    \textsuperscript{\rm 4} Meituan, China \\
    \texttt{\{zoeyyx, wnzhang\}@sjtu.edu.cn}
}

\begin{document}

\maketitle

\begin{abstract}
Agentic workflows have become the dominant paradigm for building complex AI systems, orchestrating specialized components, such as planning, reasoning, action execution, and reflection, to tackle sophisticated real-world tasks. However, systematically \emph{analyzing} and \emph{optimizing} these workflows remains challenging due to intricate component interdependencies and the lack of principled attribution methods.
In this work, we introduce \textbf{ShapleyFlow}, the first framework that employs cooperative game theory to analyze and optimize agentic workflows.
By applying the Shapley value to evaluate all possible component configurations, ShapleyFlow enables fine-grained attribution of each component's contribution and facilitates the identification of task-specific optimal configurations.
Through a constructed dataset evaluated across 7 scenarios, such as navigation, math and OS, we demonstrate 3 key contributions:
(1) \textbf{Theoretical Framework}: a principled game-theoretic approach for the attribution of contributions in agentic workflows.
(2) \textbf{Optimal Workflow Discovery}: ShapleyFlow identifies task-specific component configurations that consistently outperform workflows relying on a single LLM across all tested tasks.
(3) \textbf{Comprehensive Analysis}: we construct and analyze over 1,500 tasks, providing actionable insights and design guidelines for optimizing workflows across multiple domains.
\end{abstract}

\section{Introduction}
Large Language Models (LLMs)~\citep{brown2020language, openai2024gpt4technicalreport} have driven the rise of agentic workflows, where complex tasks are decomposed into orchestrated sequences of specialized components~\citep{sapkota2025ai, AcharyaAgenticAI}. 
Frameworks like ReAct~\citep{yao2022react} and AutoGPT~\citep{autogpt} show that structured orchestration of components can significantly improve both performance and interpretability. These workflows flexibly combine planning, reasoning, action execution, and reflection~\citep{wei2022chain,NEURIPS2023_1b44b878,Renze_2024} to address diverse challenges in areas like code generation, web navigation, and autonomous control.

Despite the widespread adoption of agentic workflows, a key challenge remains: \emph{How can we systematically analyze each component's contribution and optimize the architecture of the agentic workflow, thereby enhancing the overall performance?} Current evaluation methods~\citep{liu2023agentbench,chiang2024chatbot,guo2024stabletoolbench,yin2024mmauholisticbenchmarkagent} focus mainly on end-to-end task outcomes, without providing insights into the internal dynamics of the workflow. This black-box evaluation approach neglects the complex interdependencies between workflow components, leading to suboptimal system design and missed optimization opportunities.

There are several critical limitations in the current paradigm. First, agentic workflows simultaneously integrate multiple capabilities to solve complex tasks, where the \textbf{marginal contribution} of individual components cannot be isolated through traditional task-oriented evaluation. For example, solving a mathematical problem may require coordinated planning for strategy selection, reasoning for logical inference, and action execution for tool usage, each contributing differently depending on the specific workflow configuration. Second, vanilla ablation methods fail to capture the \textbf{synergistic effects} that emerge when workflow components interact, missing crucial insights about optimal combinations by only examining components in isolation. Third, task-specific success rates offer limited guidance for workflow optimization, making it difficult to identify which components to prioritize or how to allocate computational resources effectively.

\begin{figure*}[t]
    \centering
    \includegraphics[width=\linewidth]{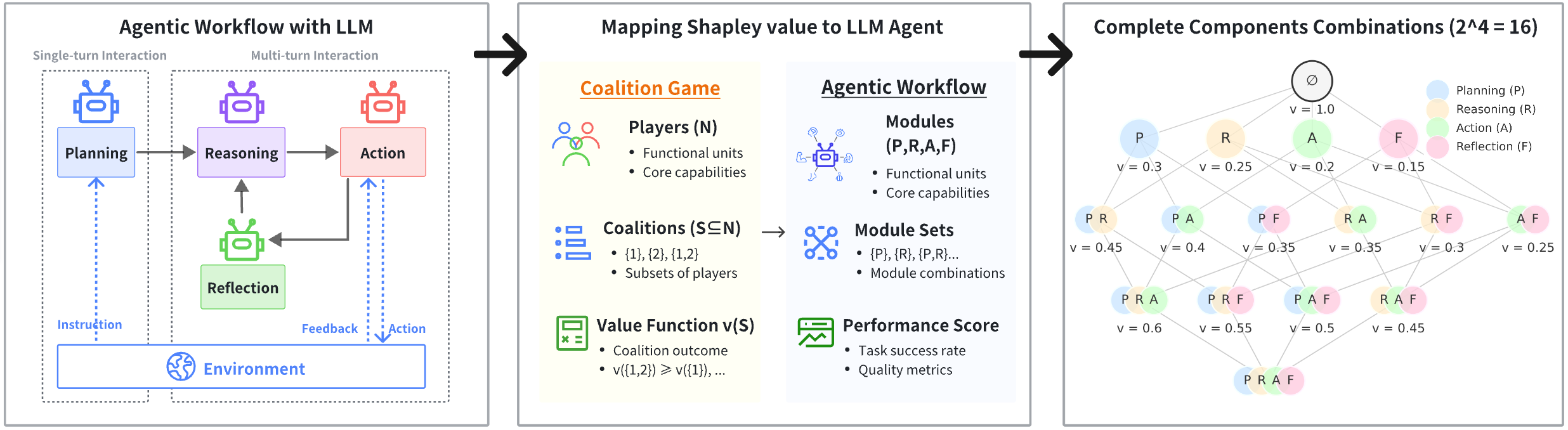}
    \vspace{-0.4cm}
    \caption{ShapleyFlow framework for agentic workflow analysis and optimization. The left panel shows a typical agentic workflow with four core components (Planning, Reasoning, Action, Reflection) orchestrated through single-turn and multi-turn interactions. The middle panel illustrates our game-theoretic formulation, where workflow components are modeled as cooperative players with coalition outcomes mapped to performance scores. The right panel demonstrates exhaustive evaluation across all possible workflow configurations ($2^4 = 16$), enabling Shapley value computation for principled component attribution and optimization guidance.}
    \label{fig:SV illustration}
    \vspace{-0.2cm}
\end{figure*}

To address these challenges, we introduce \textbf{ShapleyFlow}, an explainable framework that applies cooperative game theory to analyze and optimize agentic workflows.
Specifically, we use the Shapley value~\citep{P-295,Hart1989,10.1162/neco.2007.19.7.1939,lundberg2017unifiedapproachinterpretingmodel,pmlr-v97-ghorbani19c}, a mathematically rigorous method to quantify both individual component contributions and their interaction effects across all possible component configurations. By modeling workflow components as players in a cooperative game, ShapleyFlow enables principled attribution that shows not only what each component contributes, but also how different component combinations can be optimized for specific tasks.

This game-theoretic formulation offers several key benefits: (1) it enables fine-grained analysis of how components contribute to workflow performance under varying orchestration patterns; (2) it captures non-linear synergy effects between components that traditional methods miss; (3) it provides predictive insights for optimizing workflow configurations; and (4) it supports systematic workflow design by identifying high-impact component combinations for different task types. To our knowledge, \textbf{ShapleyFlow is the first framework to apply Shapley value theory for systematic analysis and optimization of agentic workflows}.

To demonstrate its effectiveness, we construct over \textbf{1,500 diverse agentic workflows} across diverse domains including shopping, navigation, ticket ordering, operation system, robotic control, mathematical solver, and automatic theorem proving. Each scenario requires coordinated integration of multiple workflow components, reflecting real-world demands for sophisticated agentic systems. Our ShapleyFlow analysis presents fundamental workflow design principles, identifies optimal component configurations for different tasks, and provides optimization strategies that significantly improve performance across all evaluated domains. 

We summarize our key contributions as follows:
\begin{itemize}[leftmargin=2.0em] \setlength{\itemsep}{-0.1em}
    \item \textbf{Theoretical Framework:} We introduce \emph{ShapleyFlow}, the first principled approach for analyzing and optimizing agentic workflows using cooperative game theory, enabling both component attribution and systematic workflow design.
    \item \textbf{Optimal Workflow Discovery:} We identify \emph{task-specific optimal configurations} of agentic workflows that consistently outperform any single-LLM-based workflow across diverse task categories.
    \item \textbf{Comprehensive Analysis:} We construct and analyze 1,500+ diverse tasks, providing actionable optimization guidelines for systematic workflow design across domains.
\end{itemize}

\vspace{-0.4cm}
\section{Related Work}
\subsection{Agentic Workflow Systems}
The emergence of agentic workflows has transformed AI system design from monolithic models to orchestrated multi-component architectures. Early pioneering work such as ReAct~\citep{yao2022react} demonstrated the effectiveness of structured reasoning-action cycles, establishing the foundation for component-based workflow design. AutoGPT~\citep{autogpt} advanced this paradigm by introducing autonomous workflow execution through iterative planning, tool usage, and self-reflection. MetaGPT~\citep{Hong2023MetaGPTMP} further refined workflow orchestration with hierarchical planning mechanisms that enable recursive task decomposition and role-based component coordination.

These frameworks have established the utility of modular workflow designs but primarily focus on system \emph{construction} and \emph{execution} rather than systematic \emph{analysis} and \emph{optimization}. While they demonstrate effective workflow patterns, they provide limited insights into component contribution attribution or optimization strategies for different task requirements. In contrast, our work introduces the first principled approach for analyzing component contributions within agentic workflows using cooperative game theory, enabling both fine-grained attribution and systematic optimization guidance for workflow design.

\vspace{-0.2cm}
\subsection{Workflow Analysis and Evaluation}
Evaluation frameworks for agentic systems have evolved from task-centric benchmarks to more sophisticated capability assessments. AgentBench~\citep{liu2023agentbench} provided foundational evaluation across diverse scenarios including web navigation and knowledge reasoning, but emphasized end-to-end task success without isolating underlying component contributions. MMAU~\citep{yin2024mmauholisticbenchmarkagent} introduced capability-oriented evaluation across multiple skill dimensions, yet directly maps capabilities to specific tasks, making it challenging to disentangle component-level effects from task complexity and optimize workflow configurations.

Recent benchmarks have expanded evaluation scope: OmniACT~\citep{zhang2024omniact} enables desktop environment interaction assessment, while AgentQuest~\citep{yang2024agentquest} investigates adaptive learning capabilities. However, these approaches remain fundamentally limited to black-box evaluation, providing minimal guidance for workflow optimization or component attribution. \textbf{ShapleyFlow} addresses this critical gap by applying cooperative game theory to quantitatively attribute performance across workflow components, enabling both principled component analysis and actionable optimization insights. This represents a paradigm shift from evaluation-only frameworks toward comprehensive workflow analysis and optimization methodology.

\section{ShapleyFlow: Agentic Workflow Analysis Framework}
We introduce ShapleyFlow, a framework for analyzing and optimizing workflows in agentic systems using cooperative game theory. Using the Shapley value provides a principled approach to attributing performance to components and understanding their interactions.

\subsection{Preliminaries}
\paragraph{Shapley value Definition}
The Shapley value, introduced by~\citep{P-295,Hart1989,10.1162/neco.2007.19.7.1939,lundberg2017unifiedapproachinterpretingmodel,pmlr-v97-ghorbani19c}, provides a solution concept for cooperative games that fairly distributes the rewards among players according to their marginal contributions. For a cooperative game $G = (N, v)$ with a set of players $N$ and a characteristic function $v: 2^N \rightarrow \mathbb{R}$, the Shapley value $\phi_i(v)$ for the player $i \in N$ is defined as:
\begin{equation}
\phi_i(v) = \sum_{S \subseteq N \setminus {i}} \frac{|S|!(|N|-|S|-1)!}{|N|!} \left[v(S \cup {i}) - v(S)\right]
\end{equation}
The marginal contribution $v(S \cup {i}) - v(S)$ quantifies the performance improvement when the player $i$ joins the coalition $S$, while the combinatorial weight ensures a fair averaging in all possible coalition orderings. The Shapley value satisfies key axioms: efficiency (contributions sum to total performance), symmetry (identical players receive equal attribution), dummy (non-contributing players receive zero attribution), and additivity (linear composition of games).

\subsection{Game-Theoretic Formulation}
We formulate agentic workflow analysis as a cooperative game in which workflow components collaborate to achieve optimal task performance. This game-theoretic perspective enables the principled attribution of component contributions while capturing the complex interdependencies that characterize sophisticated agentic systems.

\paragraph{Cooperative Game Definition}
We define the workflow analysis problem as a cooperative game $G = (N, v)$ where:
\begin{itemize}[leftmargin=1.5em, itemsep=0em]
\item $N$ represents the set of players (workflow components)
\item $v: 2^N \rightarrow \mathbb{R}$ is the characteristic function mapping component coalitions to performance values
\end{itemize}
For any coalition $S \subseteq N$, the characteristic function $v(S)$ measures the performance of the workflow when only the components of $S$ are active:
\begin{equation}
v(S) = \frac{1}{|T|} \sum_{t \in T} \mathbb{I}[\text{success}(t, S)]
\end{equation}
where $T$ represents the set of tasks and $\mathbb{I}[\cdot]$ is the indicator function for the success of task completion.
This formulation satisfies the essential properties of cooperative games: $v(\emptyset) = 0$ (no components yield no performance) and monotonicity $v(S) \leq v(T)$ for $S \subseteq T$ (additional components cannot decrease performance).

\paragraph{Contribution Attribution Framework}
The formulation of cooperative games naturally captures the collaborative nature of agentic workflows, where components must work together to achieve complex objectives. This formulation is particularly well-suited for analyzing workflows where component interactions are non-additive, going beyond traditional assumptions that treat components as independent building blocks with additive effects.

Our Shapley-based contribution attribution systematically evaluates all possible component combinations to understand how different orchestration patterns contribute to overall system performance. Unlike traditional ablation-based attribution that examines components in isolation by removing them one at a time, this approach provides a principled framework for capturing component interdependencies and synergistic effects.

\paragraph{Synergistic Effects Analysis}
The Shapley value framework captures synergistic relationships between workflow components that traditional evaluation methods miss. For components $i, j \in N$, we define the pairwise synergy coefficient as:
\begin{equation}
\sigma_{ij} = v({i,j}) - v({i}) - v({j}) + v(\emptyset)
\end{equation}
Positive synergy ($\sigma_{ij} > 0$) indicates complementary components that perform better together, while negative synergy suggests redundant or competing functionality.

\begin{figure}[t]
\centering
\includegraphics[width=\linewidth]{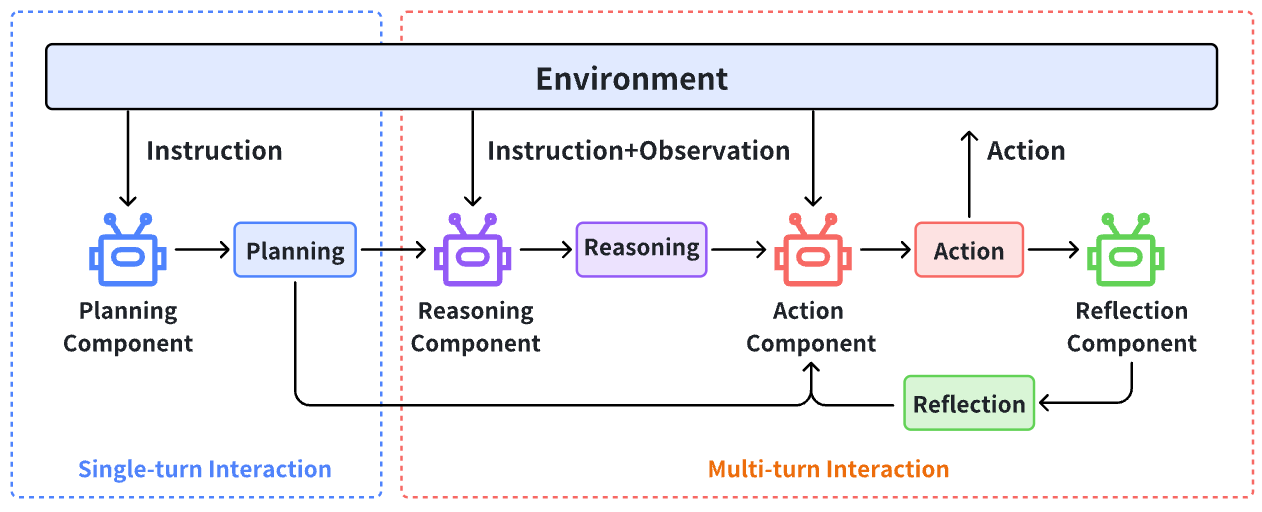}
\vspace{-0.4cm}
\caption{A vanilla agentic workflow with 4 components.}
\label{fig:Agent Workflow}
\vspace{-0.4cm}
\end{figure}

\subsection{ShapleyFlow Algorithm}

\paragraph{Agentic Workflow Design}
To demonstrate the applicability of our framework, we present a concrete instantiation using a four-component agentic architecture that captures the fundamental operational patterns observed across diverse agentic systems. Our ShapleyFlow framework is \textbf{architecture-agnostic} and can be applied to any workflow decomposition, including more fine-grained component structures.

\begin{table*}[t]
\centering
\caption{CapaBench Dataset Statistics}
\vspace{-0.2cm}
\label{tab:dataset_counts_transposed}
\resizebox{\textwidth}{!}{%
\begin{tabular}{l 
                >{\centering\arraybackslash}p{1.5cm} 
                >{\centering\arraybackslash}p{1.5cm} 
                >{\centering\arraybackslash}p{1.5cm} 
                >{\centering\arraybackslash}p{1.5cm} 
                >{\centering\arraybackslash}p{1.5cm}
                >{\centering\arraybackslash}p{1.5cm} 
                >{\centering\arraybackslash}p{1.5cm}
                >{\centering\arraybackslash}p{1.5cm} 
                >{\centering\arraybackslash}p{1.5cm} 
                >{\centering\arraybackslash}p{1.5cm} 
                >{\centering\arraybackslash}p{1.5cm} }
\toprule
\textbf{Category} & \multicolumn{2}{c}{\textbf{Shopping}}  & \textbf{Navigation} & \textbf{Ticket} & \multicolumn{2}{c}
{\textbf{Math}}& \multicolumn{3}{c}{\textbf{ATP}} & \textbf{RobotCoop} & \textbf{OS} \\
\cmidrule(lr){2-3} \cmidrule(lr){4-4} \cmidrule(lr){5-5} \cmidrule(lr){6-7} \cmidrule(lr){8-10} \cmidrule(lr){11-11} \cmidrule(lr){12-12}
\textbf{Subcategory} & Black & White & None & None & Algebra & Geometry & Coq & Lean4 & Isabelle & None & None \\
\midrule
\textbf{Count} & 48 & 62 & 250 & 150 & 250 & 250 & 111 & 111 & 111 & 100 & 102 \\
\bottomrule
\end{tabular}}
\vspace{-0.3cm}
\end{table*}

\begin{algorithm}[t]
\caption{ShapleyFlow}
\label{alg:shapleyflow}
\begin{algorithmic}[1]
\STATE \textbf{Input:} Baseline model, Target model, Task set $T$, Component set $N$
\STATE \textbf{Output:} Shapley value $\phi_i(v)$ and optimization insights
\STATE Initialize components in $N$ with baseline model
\FORALL{configuration $S \subseteq N$}
\STATE Replace components in $S$ with target model implementations
\STATE Evaluate performance $v(S)$ across task set $T$
\STATE Restore baseline models for remaining configurations
\ENDFOR
\FORALL{component $i \in N$}
\STATE Compute Shapley value $\phi_i(v)$
\ENDFOR
\RETURN $\phi_i(v)$ for all components $i$
\end{algorithmic}
\end{algorithm}

We adopt four components to build this specific agentic architecture: Planning (P), Reasoning (R), Action (A), and Reflection (F) \citep{brown2020language,yao2022react,wei2022chain, autogpt,NEURIPS2023_1b44b878, Hong2023MetaGPTMP,guo2024stabletoolbench,yin2024mmauholisticbenchmarkagent}. These components are fundamental for task decomposition, logical inference, action execution, and iterative improvement, following established patterns in agentic systems literature. While real-world systems may implement these functions through more specialized sub-components, we provide a representative case that balances analytical tractability with comprehensive coverage of key agentic capabilities.

For a new task, the workflow operates as follows:
\begin{itemize}[leftmargin=2.0em]  \setlength{\itemsep}{-0.1em}
\item \textbf{Planning (P)}: initiates the process by decomposing the task into subtasks and analyzing required resources/conditions. Planning operates in single-turn mode, establishing the initial task structure.
\item \textbf{Reasoning (R)}: receives the task, current observations, and planning output to analyze the next step progression.
\item \textbf{Action (A)}: takes planning context and current reasoning analysis to generate properly formatted actions.
\item \textbf{Reflection (F)}: is triggered based on task-specific conditions to analyze encountered problems, failure causes, and improvement suggestions. This feedback is integrated into subsequent reasoning prompts.
\end{itemize}

As illustrated in Figure~\ref{fig:Agent Workflow}, Planning operates in single-turn interaction with the environment, while Reasoning, Action, and Reflection engage in multi-turn interaction patterns, consistent with mainstream agentic workflow designs. Reasoning and Action engage in multi-turn interaction following the established ReAct architecture pattern.

\paragraph{Workflow Configuration Space}
We define the component set as $C = {P, R, A, F}$ and represent a workflow configuration as any subset $S \subseteq C$.  
Rather than claiming this four-component structure as optimal, we selected it as a representative case study that captures distinct functional roles commonly found across agentic systems.  The choice balances several practical considerations: (1) With 4 components, we evaluate $2^4 = 16$ configurations, making exhaustive Shapley analysis feasible while still capturing meaningful interactions;  (2) These components represent core capabilities that appear across diverse agentic architectures, from simple ReAct agents to complex multi-agent systems;  (3) This decomposition provides sufficient complexity to reveal interesting component dependencies and synergistic effects while remaining interpretable.

\paragraph{Algorithm Implementation}
For this specific architecture, we have a 4-player cooperative game. The characteristic function $v(S)$ evaluates performance for any subset of these components.
Algorithm~\ref{alg:shapleyflow} presents our systematic methodology for component analysis using Shapley value. This approach generalizes beyond our four-component case to any workflow architecture decomposition.

\begin{table}[t]
\centering
\setlength{\tabcolsep}{1pt}
\renewcommand{\arraystretch}{1.8}
\caption{Component Capability Coverage in CapaBench}
\label{tab:dataset_capability}
\resizebox{0.48\textwidth}{!}{
\begin{tabular}{p{0.5cm} p{3cm} c c c c c c c c c}
\specialrule{1pt}{0pt}{0pt} 
\multicolumn{2}{l}{\multirow{2}{*}{\textbf{}}} & \multicolumn{3}{c}{\textbf{Daily Activities}} & \multicolumn{3}{c}{\textbf{Computation}} & \multicolumn{1}{c}{\textbf{Role Control}} \\
\cline{3-5}
\cline{6-8}
\cline{9-9}
\multicolumn{2}{c}{} & \textbf{Shopping} & \textbf{Navigation} & \textbf{Ticket} & \textbf{Math} & \textbf{ATP} & \textbf{OS} & \textbf{RobotCoop} \\
\hline
\multirow{2}{*}{\textbf{P}} & Task Steps & {$\checkmark$} & & & {$\checkmark$} & {$\checkmark$} & &  \\
& Resource Constraints & & {$\checkmark$} & {$\checkmark$} & & & {$\checkmark$} & {$\checkmark$} \\
\hline
\multirow{2}{*}{\textbf{R}} & Logical Validation & & & & {$\checkmark$} & {$\checkmark$} & {$\checkmark$} & \\
& Knowledge Inference & {$\checkmark$} & {$\checkmark$} & {$\checkmark$} & & & & {$\checkmark$} \\

\hline
    \textbf{A} & Environment Actions  & & & & $\checkmark$ & $\checkmark$ & $\checkmark$ &  \\
& Interactive Actions & \textbf{$\checkmark$} & \textbf{$\checkmark$} & \textbf{$\checkmark$} & & &
&$\checkmark$ \\

\hline
\textbf{R} & Failure Analysis & {$\checkmark$}& {$\checkmark$} & {$\checkmark$} & {$\checkmark$} & {$\checkmark$} & {$\checkmark$} & {$\checkmark$} \\
\specialrule{1pt}{0pt}{0pt} 
\end{tabular}}
\vspace{-0.2cm}
\end{table}

\subsection{Benchmark Construction}
To demonstrate the effectiveness of our ShapleyFlow framework, we construct \textbf{CapaBench} (\textbf{Capa}bility-level Assessment \textbf{Bench}mark), a comprehensive benchmark containing over 1,500 agentic workflows spanning diverse domains. Unlike existing benchmarks that focus on end-to-end evaluation, CapaBench is specifically designed to enable systematic component analysis through our game-theoretic framework.

Our workflow construction follows three key principles: (1) \textbf{Component Integration}: Each workflow requires meaningful coordination between Planning, Reasoning, Action, and Reflection components;  (2) \textbf{Realistic Complexity}: Workflows mirror real-world agentic system demands across different domains;  (3) \textbf{Capability Coverage}: Tasks span the full spectrum of component capabilities as shown in Table~\ref{tab:dataset_capability}.

\input{Table/mainResult}

CapaBench comprises seven task categories organized into three categories:
\begin{itemize}[leftmargin=2.0em]  \setlength{\itemsep}{-0.1em}
 \item \textbf{Daily Activities}: Online Shopping (Shopping), Navigation Planning (Navigation), Ticket Ordering (Ticket).
 \item \textbf{Computation}: Mathematical Solver (Math), Automatic Theorem Proving (ATP), Operating System (OS).
\item \textbf{Role Control}: Robot Coordination (RobotCoop).
\end{itemize}

This taxonomy enables systematic analysis of how optimal workflow configurations vary across different operational contexts.
Table~\ref{tab:dataset_counts_transposed} shows the distribution of 1,535 total tasks across categories. Detailed construction methodology is provided in supplementary material.

\begin{figure*}[t]
    \centering
    \includegraphics[width=\linewidth]{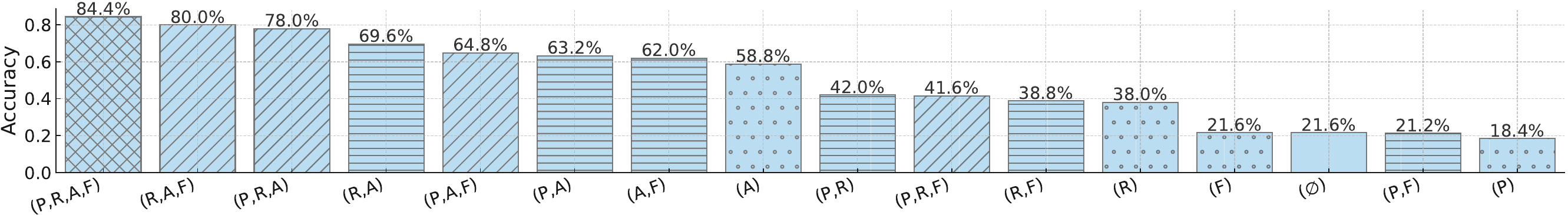}
    \vspace{-0.7cm}
    \caption{Results of all combinations in Math (Algebra) for Claude-3.5-Sonnet under different configurations. The pattern of the bars indicates the number of components (ranging from 0 to 4) that Claude is involved in.}
    \label{fig:Model Configuration Accuracy Comparison}
    \vspace{-0.2cm}
\end{figure*}

\begin{figure*}[t]
    \centering
    \begin{minipage}[t]{0.235\textwidth}
        \centering
        \includegraphics[width=\textwidth]{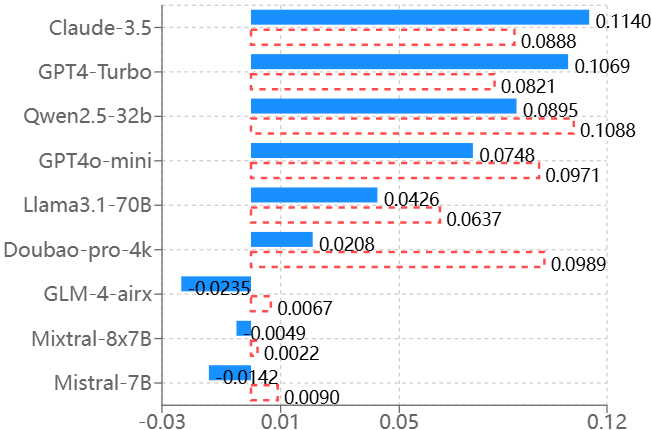} 
        \subcaption{Planning}
    \end{minipage}
    \hfill
    \begin{minipage}[t]{0.235\textwidth}
        \centering
        \includegraphics[width=\textwidth]{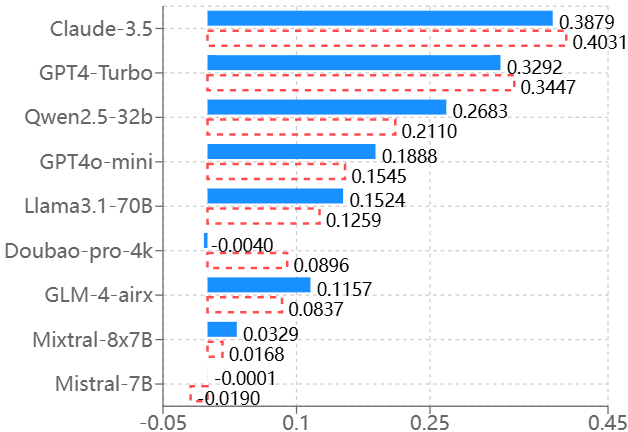}
        \subcaption{Reasoning}
    \end{minipage}
    \hfill
    \begin{minipage}[t]{0.245\textwidth}
        \centering
        \includegraphics[width=\textwidth]{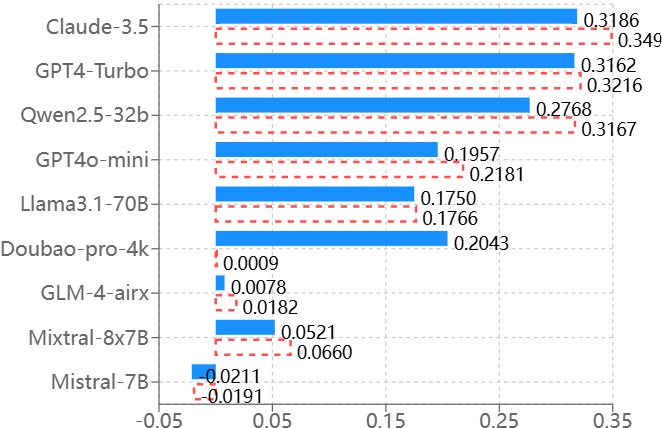} 
        \subcaption{Action}
    \end{minipage}
    \hfill
    \begin{minipage}[t]{0.235\textwidth}
        \centering
        \includegraphics[width=\textwidth]{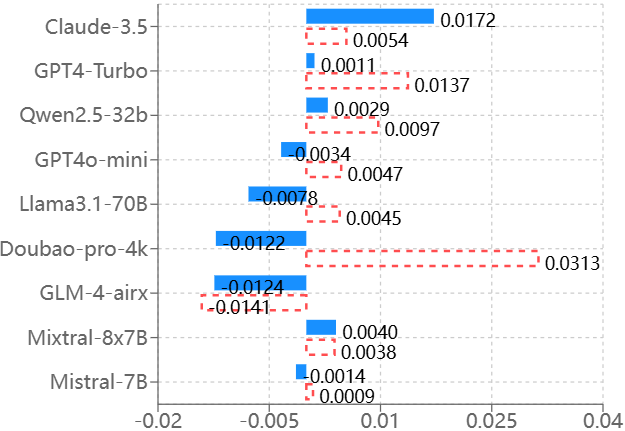} 
        \subcaption{Reflection}
    \end{minipage}

    \includegraphics[width=0.3\textwidth]{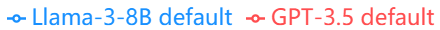} 
    \vspace{-0.2cm}     
    \caption{Comparation of Shapley value under different default models.}
    \label{fig:ablation study}
\vspace{-0.3cm}   
\end{figure*}

\section{Experiments}
\subsection{Experimental Setup}
We apply ShapleyFlow to analyze component contributions across 1,500+ agentic workflows using systematic configuration testing. For each workflow, we employ \textbf{Llama3-8B-Instruct} as the baseline implementation for all components (Planning, Reasoning, Action, Reflection), then systematically replace components with target implementations to evaluate all $2^4 = 16$ possible configurations. This approach isolates the contribution of each component upgrade while maintaining consistent baseline conditions.

The choice of \textbf{Llama3-8B-Instruct} as the default model implementation is motivated by three factors:
(1) it is open-source and easy to deploy at scale, making it practical for large-scale experiments;
(2) its lightweight architecture ensures efficient evaluation of thousands of workflows;
(3) its moderate task success rates create a balanced baseline, allowing the performance impact of replacing components with more advanced models to be clearly observed and quantified.

We evaluate 9 representative LLMs across three categories:
\begin{itemize}[leftmargin=2.0em]  \setlength{\itemsep}{-0.1em}
 \item \textbf{Closed API Models}: Claude-3.5-Sonnet~\citep{anthropic2024claude35}, GPT-4-turbo~\citep{openai2023gpt4turbo}, GPT-4o-mini~\citep{openai2024gpt4o}, GLM-4-air~\citep{thudm2024glm4air}, Doubao-pro-4k~\citep{doubao2024}.
 \item \textbf{Mid-parameter Open-Source Models (40B-100B)}: Llama3.1-70B-Instruct~\citep{llama2024}, Mixtral-8x7B-Instruct (46.7B)~\citep{mistral87}.
\item \textbf{Low-parameter Open-Source Models ($\leq$32B)}: Qwen2.5-32B-Instruct~\citep{alibaba2024qwen25}, Mistral-8B-Instruct~\citep{mistral2023mistral8b}.
\end{itemize}

All experiments use consistent settings (temperature=0.0, max tokens=2048, top-p=1.0) on NVIDIA A100-80GB GPUs with vLLM for efficient inference. The evaluation metric is task success rate, measuring the proportion of successfully completed tasks across all tasks (more details are provided in the supplementary material).

\subsection{Component Contribution Attribution}
Table~\ref{tab:consolidated_results_vertical} presents comprehensive component contribution analysis across diverse workflow categories. The Planning (P), Reasoning (R), Action (A), and Reflection (F) rows show individual component Shapley value $\phi_i(v)$, quantifying the marginal contribution of replacing the baseline Llama3-8B-Instruct implementation with the target model for component $i$ in each workflow domain. 

\subsection{Optimal Configuration Discovery}
Results marked with '*' below each dataset indicate the optimal workflow configurations where each component utilizes the LLM with the highest Shapley value for that component. Leveraging high-Shapley components consistently maximizes performance: e-commerce workflows achieve 43.31\% optimal accuracy, while formal verification workflows reach 86.79\%. These results demonstrate ShapleyFlow's ability to predict and recommend optimal component combinations, enabling systematic workflow optimization.

\input{Figure/radarChart/radar}

\subsection{Component Impact Validation}
Figure~\ref{fig:Model Configuration Accuracy Comparison} demonstrates that our Shapley value attribution accurately predicts workflow performance. Using Claude-3.5-Sonnet on mathematical reasoning workflows, we observe clear correspondence between component contributions and configuration performance. High-Shapley configurations like (P,R,A) achieve 78.0\% success rate, dramatically outperforming the 21.6\% baseline. Incremental improvements align with predictions: adding Planning alone improves performance to 18.4\%, while the Planning-Action combination reaches 63.2\%. Low-contribution configurations like (P,F) yield poor performance (21.2\%), confirming Shapley value accurately quantify component importance.

\section{Analysis}

\paragraph{Cross-Domain Optimization Patterns}
Figure~\ref{fig:radar chart} demonstrates distinct optimization strategies across workflow categories. Computation-intensive workflows (Math, ATP-Coq) benefit most from Action component upgrades, while interactive workflows (OS, RobotCoop) show strongest gains from Reasoning-Planning combinations. This systematic analysis enables domain-specific optimization strategies, guiding practitioners toward the most effective component investments for their use cases.

\paragraph{Component Specialization Patterns}
Our comprehensive analysis shows systematic component specialization across workflow types:
\begin{itemize}  [leftmargin=2.0em]  
    \item \textbf{Reasoning-Dominant Workflows}: Interactive scenarios (Ticket, RobotCoop, OS) show highest Reasoning contributions. These workflows require dynamic decision-making, constraint balancing, and real-time adaptation. Strong Reasoning components enable effective uncertainty handling and logical inference under evolving conditions.

    \item \textbf{Action-Dominant Workflows}: Precision-critical tasks (Shopping, Math, ATP) prioritize Action components. These workflows demand exact procedural execution, syntactic correctness, and systematic verification. Robust Action components ensure reliable step-by-step execution without errors.

    \item \textbf{Reflection Component Analysis}:  Across most workflow types, Reflection components show consistently lower Shapley value, indicating limited impact on task performance. This pattern likely stems from two factors: (1) task success rates inadequately capture reflection quality, a model's ability to analyze its own mistakes doesn't directly translate to improved outcomes; (2) without guidance from more capable models, Reflection components struggle to identify error root causes, limiting their practical effectiveness in driving performance improvements.
\end{itemize}

\paragraph{Framework Robustness Analysis}
We validate framework robustness by testing sensitivity to baseline model selection. Replacing Llama3-8B-Instruct with GPT-3.5-Turbo as the baseline model, we re-analyze component contributions across Robot Cooperation workflows. Figure~\ref{fig:ablation study} shows absolute Shapley value shift with baseline strength, but relative component rankings remain highly consistent.
We use the pairwise consistency metric to quantify ranking stability:
\begin{equation}
\text{Consistency Rate} = \frac{\text{Consistent Model Pair Rankings}}{\text{Total Model Pairs}}
\end{equation}

Results show strong consistency: Reasoning (91.67\%), Action (86.11\%), Planning (72.22\%), with overall rate of 85.18\%. This demonstrates that ShapleyFlow provides reliable contribution attribution regardless of baseline choice, ensuring robust optimization recommendations.

\begin{figure}[t]
    \centering
    \includegraphics[width=0.85\linewidth]{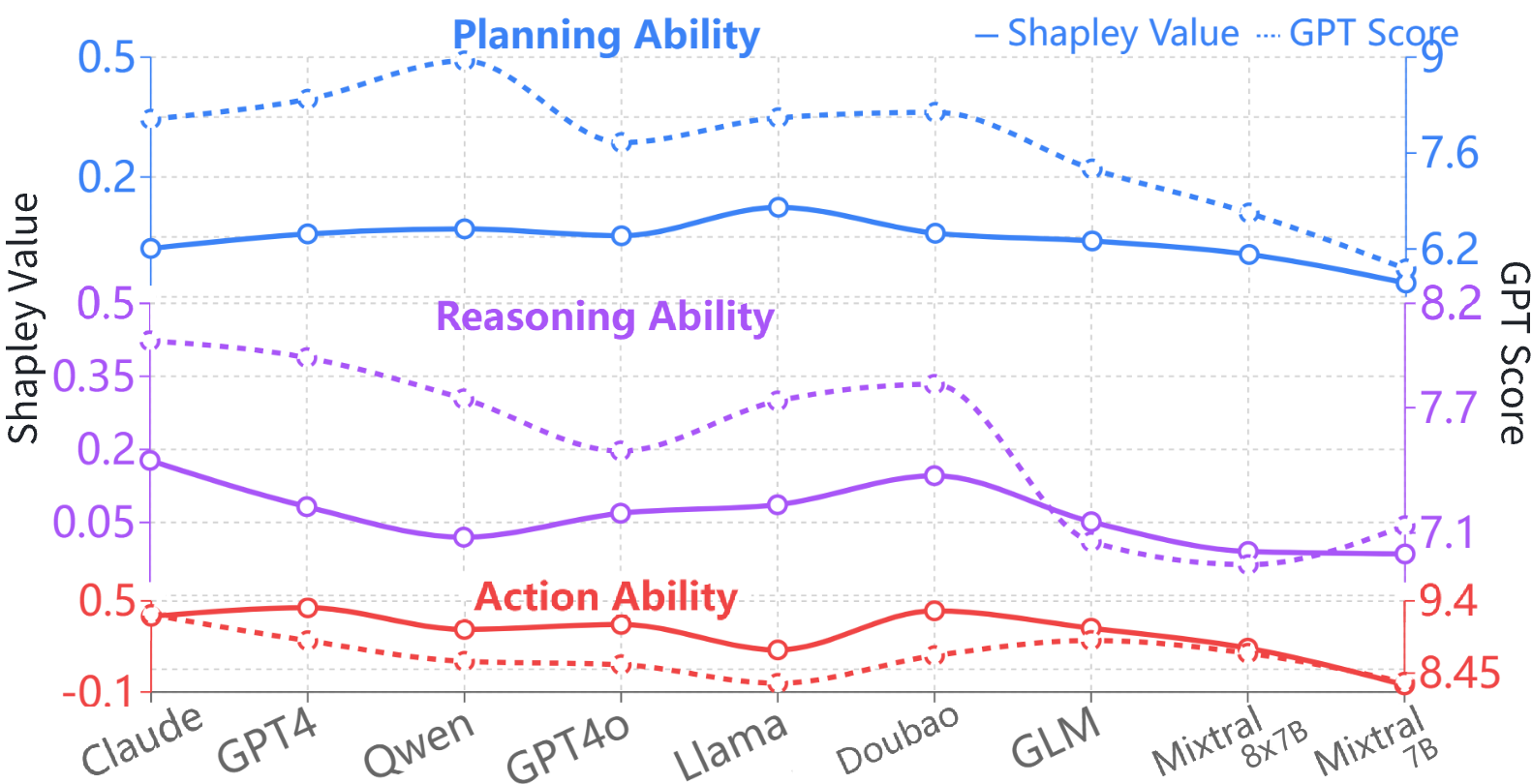}
    \vspace{-0.2cm}
    \caption{Planning, Reasoning, and Action Evaluation on Algebra. The left Y-axis shows Shapley value with solid lines and the right shows the GPT scores with dashed lines.}
    \label{fig:comp_line_chart}
    \vspace{-0.4cm}
\end{figure}

\paragraph{Attribution Consistency Validation}
To validate that ShapleyFlow's contribution attribution aligns with traditional evaluation methods, we conduct a consistency study by applying both ShapleyFlow and GPT-o1-mini based LLM-as-judge evaluation on successful Algebra workflow trajectories. We evaluate 2,180 single-step workflow samples, where GPT-o1-mini assesses semantic rationality and task completion for Planning/Reasoning components, and logical comprehension for Action components. 
Figure~\ref{fig:comp_line_chart} demonstrates strong correlation between Shapley value and independent GPT assessment scores across different components: Planning (0.81), Reasoning (0.77), Action (0.67). The consistent ranking patterns across diverse methods confirm that ShapleyFlow's contribution attribution captures component-specific capabilities that align with existing automated evaluation approaches.

\paragraph{Comparative Analysis with Ablation Studies}
Compared to traditional ablation methods requiring only $n+1$ evaluations, our $2^n$ approach demands increased computational cost (16 vs. 5 evaluations for 4 components). However, this investment yields substantially richer optimization insights by capturing component interactions that ablation studies fundamentally miss. 
For example, while ablation analysis might compare PRA and PRAF configurations to estimate Reflection's contribution, this approach overlooks Reflection's context-dependent utility. Reflection components heavily depend on Action execution and subsequent observations, comparing PR vs. PRF configurations could yield opposite conclusions about Reflection's value compared to PRA vs. PRAF, since Reflection without prior Action lacks meaningful feedback to process.
Our comprehensive evaluation demonstrates that component contributions are inherently conditional on collaboration patterns, making Shapley value's systematic consideration of all possible coalitions essential for accurate attribution. Unlike ablation studies that assume independent component effects, ShapleyFlow captures the cooperative game dynamics where component utility depends critically on the presence and performance of complementary components, enabling principled optimization decisions that account for workflow interdependencies.

\section{Conclusion}
We introduce \textbf{ShapleyFlow}, the first evaluation framework to systematically quantify component contributions in agentic workflows by using cooperative game theory. By contribution attribution to key components—planning, reasoning, action, and reflection—ShapleyFlow enables principled analysis and optimization of complex workflows. Our experiments across 9 backbone LLMs, 7 task categories, and over 1,500 workflows demonstrate the effectiveness of our approach in identifying task-specific optimal configurations.

ShapleyFlow remains tractable even as workflows grow in complexity. This can be achieved by restricting analysis to task-relevant subsets or by employing efficient approximation techniques such as KernelSHAP~\citep{lundberg2017unifiedapproachinterpretingmodel}, SVARM~\citep{kolpaczki2024approximatingshapleyvaluemarginal}, or other sampling-based estimators.

While this work focuses on the Shapley value, alternative attribution methods such as the Banzhaf Value~\citep{DRAGAN1996451} and Shapley Interactions~\citep{muschalik2024shapiqshapleyinteractionsmachine} represent promising directions for future research. These methods may offer complementary insights and enable more expressive modeling of component interactions. Future extensions of ShapleyFlow could incorporate these alternatives to enhance attribution fidelity and support automated, domain-general workflow discovery and optimization.


\bibliography{aaai2026}


\section*{Appendix for Dataset Details}
\label{sec:appendix}
\begin{table*}[t]
  \caption{PRAF Experiment Results on Mathematics Tasks with $\Delta$ Accuracy}
  \label{tab:praf_math}
  \centering
  \resizebox{\textwidth}{!}{%
  \begin{tabular}{c|ccccc|c|ccccc|c}
    \toprule
    & \multicolumn{6}{c|}{Algebra} & \multicolumn{6}{c}{Geometry} \\
    LLM & Pt & Rt & At & Ft & Acc(\%) & $\Delta$ Acc(\%) & Pt & Rt & At & Ft & Acc(\%) & $\Delta$ Acc(\%) \\
    \midrule
    \texttt{llama3-8B-instruct} & / & / & / & / & 21.6 & / & / & / & / & / & 14.4 & / \\
    \texttt{Claude-3.5-Sonnet} & 0.021 & \textbf{0.177} & 0.398 & \underline{0.031} & \underline{84.4} & 62.8& 0.055 & \textbf{0.085} & 0.486 & \textbf{0.054} & \underline{82.4} & 68.0\\
    \texttt{gpt-4-turbo} & 0.058 & 0.082 & \textbf{0.456} & 0.020 & 83.2 & 61.6& 0.038 & 0.047 & \underline{0.527} & 0.025 & 78.0 & 63.6\\
    \texttt{qwen2.5-32B} & 0.059 & \underline{0.146} & \underline{0.436} & 0.011 & \textbf{86.8} & 65.2& \underline{0.071} & \underline{0.067} & \textbf{0.530} & \underline{0.051} & \textbf{86.4} & 72.0\\
    \texttt{gpt-4o-mini} & \underline{0.070} & 0.020 & 0.313 & \textbf{0.053} & 67.2 & 45.6& 0.065 & 0.024 & 0.368 & 0.035 & 63.6 & 49.2\\
    \texttt{doubao-pro-4k} & \textbf{0.124} & 0.086 & 0.178 & 0.004 & 60.8 & 39.2& \textbf{0.105} & 0.032 & 0.186 & -0.007 & 46.0 & 31.6\\
    \texttt{GLM-4-air} & 0.053 & 0.069 & 0.346 & 0.004 & 68.8 & 47.2& 0.059 & 0.019 & 0.349 & 0.006 & 57.6 & 43.2\\
    \texttt{llama3-70B} & 0.040 & 0.051 & 0.321 & 0.007 & 63.6 & 42.0& 0.015 & 0.011 & 0.333 & 0.005 & 50.8 & 36.4\\
    \texttt{Mistral-8X7B} & 0.006 & -0.010 & 0.190 & -0.010 & 39.2 & 17.6& 0.004 & 0.016 & 0.138 & -0.018 & 28.4 & 14.0\\
    \texttt{Mistral-7B} & -0.065 & -0.015 & -0.053 & -0.003 & 8.0 & -13.6& -0.055 & 0.014 & -0.035 & -0.004 & 6.4 & -8.0\\
    \bottomrule
  \end{tabular}%
  }
\end{table*}

\begin{table*}[t]
  \caption{Experiment Results on Automatic Theorem Proving Tasks with $\Delta$ Accuracy}
  \label{tab:praf_formal}
  \centering
  \resizebox{\textwidth}{!}{%
  \begin{tabular}{c|ccccc|c|ccccc|c|ccccc|c}
    \toprule
    & \multicolumn{6}{c|}{Coq} & \multicolumn{6}{c|}{Lean 4} & \multicolumn{6}{c}{Isabelle} \\
    LLM & Pt & Rt & At & Ft & Acc(\%) & $\Delta$ Acc(\%) & Pt & Rt & At & Ft & Acc(\%) & $\Delta$ Acc(\%) & Pt & Rt & At & Ft & Acc(\%) & $\Delta$ Acc(\%) \\
    \midrule
    \texttt{llama3-8B} & / & / & / & / & 6.4 & / & / & / & / & / & 2.7 & / & / & / & / & / & 7.2 & / \\
    \texttt{Claude-3.5} & 0.010 & \textbf{0.067} & \textbf{0.795} & \underline{0.027} & \textbf{96.4} & 90.0& 0.002 & \textbf{0.059} & \textbf{0.662} & \textbf{0.098} & \textbf{84.7} & 82.0& 0.025 & \underline{0.046} & \underline{0.523} & \textbf{0.082} & \textbf{74.8} & 67.6\\
    \texttt{gpt-4-turbo} & \underline{0.032} & 0.038 & \underline{0.706} & 0.024 & \underline{86.5} & 80.1& -0.015 & -0.006 & 0.375 & 0.033 & 41.4 & 38.7& 0.020 & \textbf{0.048} & \textbf{0.542} & 0.012 & \underline{69.4} & 62.2\\
    \texttt{qwen2.5-32B} & 0.014 & 0.029 & 0.615 & 0.026 & 74.8 & 68.4& -0.007 & 0.020 & \underline{0.486} & \underline{0.050} & \underline{57.7} & 55.0& \underline{0.048} & 0.041 & 0.434 & \underline{0.036} & 63.1 & 55.9\\
    \texttt{gpt-4o-mini} & \textbf{0.038} & -0.016 & 0.391 & 0.018 & 49.5 & 43.1& -0.013 & -0.020 & 0.396 & 0.007 & 39.6 & 36.9& 0.030 & -0.012 & 0.249 & 0.021 & 36.0 & 28.8\\
    \texttt{doubao-pro-4k} & 0.007 & 0.039 & 0.204 & 0.001 & 31.5 & 25.1& -0.017 & \underline{0.029} & 0.095 & 0.028 & 16.2 & 13.5& 0.035 & 0.007 & -0.064 & 0.004 & 5.4 & -1.8\\
    \texttt{GLM-4-air} & 0.015 & 0.016 & 0.115 & \textbf{0.033} & 24.3 & 17.9& -0.004 & 0.005 & 0.193 & 0.013 & 23.4 & 20.7& -0.006 & -0.006 & 0.176 & 0.017 & 25.2 & 18.0\\
    \texttt{llama3-70B} & 0.018 & -0.137 & 0.190 & 0.009 & 14.4 & 8.0& -0.005 & -0.000 & 0.030 & 0.020 & 7.2 & 4.5& 0.043 & -0.032 & 0.155 & 0.005 & 24.3 & 17.1\\
    \texttt{Mistral-8X7B} & 0.014 & \underline{0.056} & 0.122 & 0.014 & 27.0 & 20.6& \underline{0.003} & -0.017 & 0.068 & -0.018 & 6.3 & 3.6& \textbf{0.058} & 0.014 & -0.071 & -0.028 & 4.5 & -2.7\\
    \texttt{Mistral-7B} & 0.018 & 0.013 & 0.028 & -0.015 & 10.8 & 4.4& \textbf{0.020} & 0.011 & 0.012 & 0.012 & 8.1 & 5.4& -0.014 & 0.006 & -0.068 & 0.003 & 0.0 & -7.2\\
    \midrule
    \texttt{best} & / & / & / & / & \textit{94.6} & \textit{+88.2} & / & / & / & / & \textbf{\textit{87.4}} & \textit{+84.7} & / & / & / & / & \textbf{\textit{78.4}} & \textit{+71.2} \\
    \bottomrule
  \end{tabular}}
\end{table*}

\subsubsection*{Online Shopping}
The Online Shopping dataset is designed to evaluate agents' planning, reasoning, and action capabilities in completing e-commerce tasks. The dataset consists of \textbf{110 tasks}, divided into two parts: \textbf{white-box tasks (62)}, which are from the Webshop dataset, and \textbf{black-box tasks (48)}, which are expanded using GPT-4 to enhance instruction diversity and complexity.

Dataset expansion was constructed by modifying instructions from the original dataset. GPT-4 was used to rephrase instructions for greater linguistic diversity, adding context or background such as \textit{“Next week is Halloween, and I need themed decorations.”} Additionally, parameters were enriched with attributes like size, color, or material to increase task complexity. For challenging cases, explicit prompts were created to guide planning, for example, \textit{“First search for desks with wood finishes, then filter by size and price.”}

A typical instruction in Online Shopping might be:
\textit{“I'm looking for a small portable folding desk that is already fully assembled; it should have a khaki wood finish, and price lower than 140 dollars, and length bigger than 40 inches.”}

Agents are evaluated based on their ability to follow optimal trajectories, such as:
\begin{itemize}
    \item \textbf{Ideal Trajectory 1:} Search for all attributes directly \textit{("desk, wood, folding, khaki, 40 inches, \$140")} and proceed to the target item.
    \item \textbf{Ideal Trajectory 2:} Broad search \textit{("desk, wood, folding")}, filter by price, and then refine attributes (color, size).
\end{itemize}

\subsubsection*{Navigation Planning}
\begin{figure*}[t]
    \centering
    \includegraphics[width=0.95\linewidth]{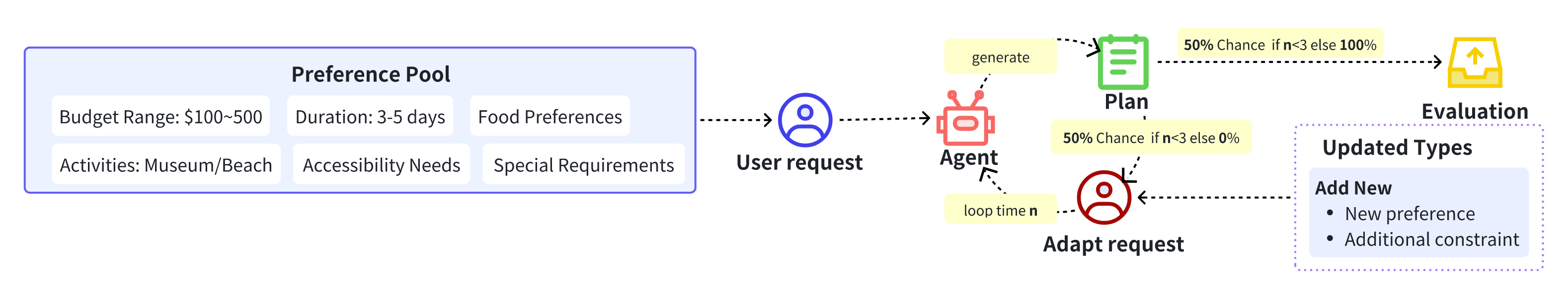}
    \vspace{-0.2cm}
    \caption{An Example Problem in Three Languages.}
    \label{fig:navigation_planning_planning}
\end{figure*}

The Navigation Planning dataset assesses collaborative itinerary generation with dynamic constraint adaptation, containing \textbf{250 tasks} developed through enhanced automated generation. As shown in Figure~\ref{fig:navigation_planning_planning}, our framework extends \citep{lin2023decision} with three key innovations:

\begin{itemize}
    \item \textbf{Precision Evolution}: Each task begins with three core requirements (e.g., "\$3,000 budget for 4 adults"), with 50\% probability per interaction round to introduce new constraints from predefined pools (accessibility needs, seasonal activities). 

    \item \textbf{Location Profiling}: We implement stochastic sampling of destination attributes: \begin{itemize} \item \textit{Accessibility}: Transportation options (train/bus connectivity) \item \textit{Amenities}: Family/pet-friendly facilities \item \textit{Pricing}: Seasonal price fluctuations (±15\%) \end{itemize}
    
    \item \textbf{Evaluation Protocol}:We evaluates the rationality of the planned route, based on how well the proposal aligns with user preferences, considering factors such as budget adherence, inclusion of specified activities, and efficient travel distances.  
\end{itemize}

A sample task evolves from initial requirements "\textit{7-day Japan tour under \$4k}" to include "\textit{must visit at least two UNESCO sites}" during planning. Agents must preserve previous constraints while integrating new ones, testing sequential reasoning capabilities. Our automated validator ensures solution feasibility through geographic coordinate verification and budget accounting simulations.

\subsection{Ticket Ordering.}
The Ticket Ordering task evaluates the ability of agents to collaboratively provide the best flight combinations for two users. The dataset consists of 150 tasks, which are designed to benchmark the performance of different agents in ticket ordering.

Inspired by the framework presented by \citep{lin2023decision}, we build our evaluation framework based on their structure. Specifically, we use the provided code to generate the dataset, which includes two users' calendars. The tasks are created by combining the users' calendar data, and agents are then asked to provide flight recommendations based on this information.



Table \ref{tab:consolidated_results_vertical} shows the experimental results for the Ticket Ordering task.
The baseline model achieves an accuracy of 19.94\%. \texttt{Claude-3.5-Sonnet} achieves the highest accuracy of 62.85\%, improving by +42.91\%. \texttt{gpt-4-turbo-0409} follows with an accuracy of 54.37\%, improving by +34.43\%. The accuracy range, from 0.0\% (\texttt{Mistral-7B-Instruct}) to 62.85\%, highlights the dataset's ability to differentiate models based on their performance.

The dataset emphasizes Reasoning and Action capabilities, as seen in the high \texttt{R} and \texttt{A} Shapley values for top models like \texttt{Claude-3.5-Sonnet}, \texttt{gpt-4-turbo-0409}, and \texttt{qwen2.5-32b-Instruct}. Models with stronger Reasoning and Action abilities show significant accuracy improvements, whereas those with lower values for these modules, such as \texttt{Mistral-7B-Instruct}, experience considerable performance deficits.






\subsection*{Math Solver}
The Math Solver dataset evaluates agents' planning, reasoning, and action capabilities in solving diverse mathematical problems, with a particular focus on tool usage during the problem-solving process. This dataset is divided into two categories: \textbf{Algebra} and \textbf{Geometry}, comprising a total of \textbf{500 tasks} (\textbf{250 Algebra tasks} and \textbf{250 Geometry tasks}).

\textbf{Dataset Construction.} The dataset is derived from the MATH dataset and enhanced with GPT-4 to improve diversity and relevance. The MATH dataset's original structure includes a large number of highly similar questions without detailed knowledge point categorization, making evaluation costly and inefficient. To address this, we synthesized new data by:
\begin{itemize}
    \item [(1)]Summarizing Knowledge Points: All problems in the MATH dataset were analyzed using GPT-4 to extract a comprehensive list of key concepts.
    \item [(2)]Condensing Categories: GPT-4 distilled the extracted concepts into \textbf{10 key knowledge points} for Algebra and Geometry, respectively.
    \item [(3)]Mapping Labels: Each problem in the original dataset was mapped to one of the 10 knowledge points and assigned a difficulty level (1–5).
    \item [(4)]Synthesizing New Problems: For each unique combination of knowledge point and difficulty level, GPT-4 generated five new problems, ensuring coverage across all categories.
\end{itemize}
Overall, both algebra and geometry each include ten knowledge points. Each knowledge point is divided into five levels, and for each combination, there are five problems. Therefore, the total amount of data is $2 \times 10 \times 5 \times 5 = 500$.
Knowledge points and corresponding examples can be seen in Table.\ref{tab:konwledge_point_math}.

\subsection*{Automatic Theorem Proving}
\begin{figure*}[h]
    \centering
    \includegraphics[width=0.95\linewidth]{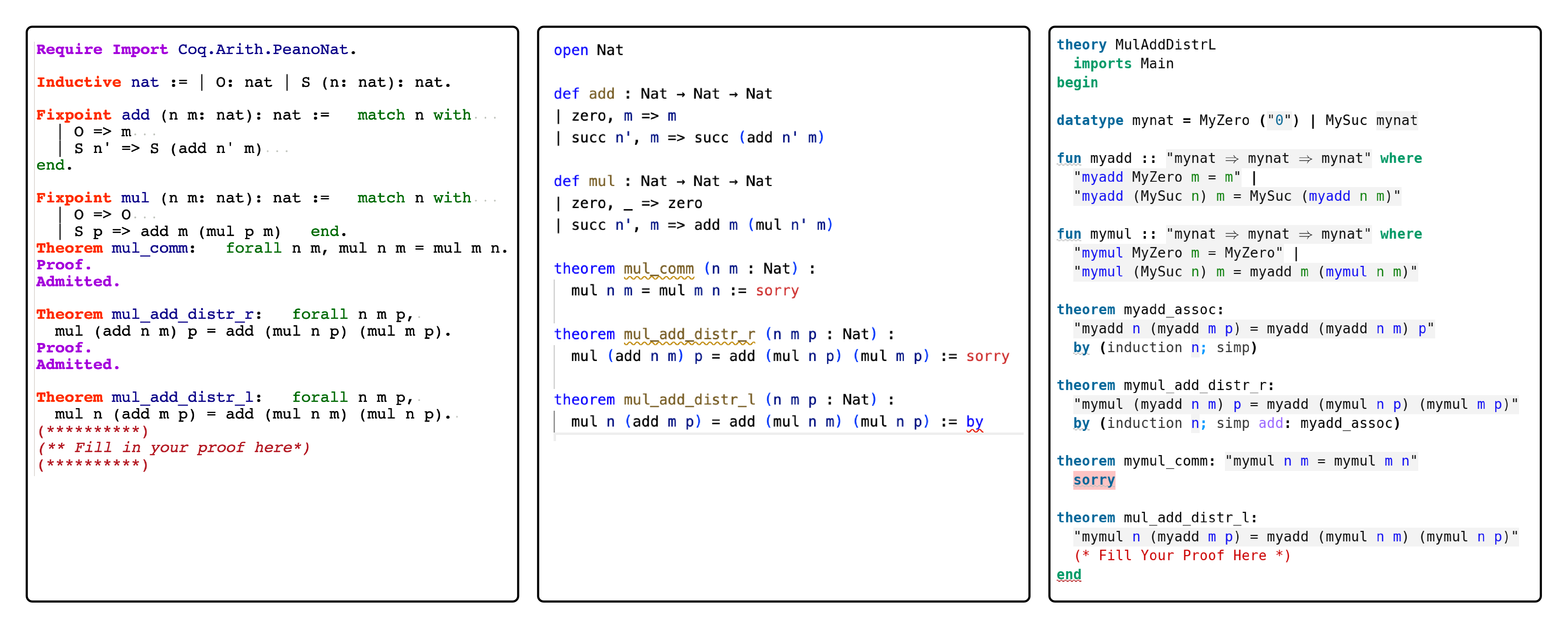}
    \vspace{-0.2cm}
    \caption{An Example Problem in Three Languages.}
    \label{fig:coq_lean_isabelle}
\end{figure*}
The Automatic Theorem Proving dataset evaluates agents' capabilities in solving formal proof problems, focusing on generating code for logical proofs. The dataset includes three categories: \textbf{Coq}, \textbf{Lean 4}, and \textbf{Isabelle}, with a total of \textbf{333 tasks} (111 tasks per category).

\textbf{Dataset Construction.}  The dataset originates from 111 Coq problems curated from course material, covering the following topics:

\begin{itemize}
    \item [(1)] Algebraic Calculations, e.g., derivation of linear systems.
    \item [(2)] Properties of Functions, e.g., translation and monotonicity of functions.
    \item [(3)] Properties of Recursive Structures, e.g., operations on tree structures.
    \item [(4)] Logical Problems, e.g., relationships between AND, OR, and NOT.
    \item [(5)] Properties of Natural Numbers, e.g., proving 6 is not a prime number.
\end{itemize}

These proof problems serve as introductory exercises in college formal proof courses, focusing on basic syntax and simple logical relationships. They are challenging for students, making them a suitable benchmark for evaluating the performance of LLMs.

To comprehensively assess LLMs' formal proof capabilities, these problems were further translated into Lean 4 and Isabelle versions. Coq, Lean 4, and Isabelle are widely used formal proof languages, and using multiple languages allows for a more rigorous comparison of model capabilities.

\subsection*{Operating System}
The Operating System dataset evaluates an agent's ability to interact with a simulated OS terminal by executing commands to address OS-related tasks, comprising 71 Ubuntu terminal tasks and 31 Git tasks.

In Ubuntu tasks, agents are required to propose bash commands to execute in Ubuntu Terminal and get feedback from the terminal to complete the task. We utilized the AgentBench-OS framework to employ the evaluation.

We enhanced the automated data generation method from AgentBench-OS to construct our new dataset, primarily generating operation-type data. The original method leverages LLMs to generate tasks and employs unit tests to ensure their accuracy. While creating the dataset, we used specific prompts to guide the generation of desired data types. The dataset comprises 71 AgentBench-OS tasks, categorized into 40 file system manipulation, 20 system setting, and 11 process running tasks.

For the git tasks, we selected data from learngitbranching. The learngitbranching website itself is a tutorial git beginner. It provides terminal and sandbox environment that simulates git using a tree structure. Git tree dynamically updates along with each git command from the terminal. Given initial and target states for both local and remote git trees, agents must interact with the git tree via the terminal to transform it from its initial state to the target state. The dataset assesses proficiency in fundamental git commands and their combination to execute advanced git functionalities.

\begin{table*}[htbp]
  \caption{Categories and Examples of Operating System Datasets}
  \label{tab:konwledge_point_math}
  \centering
  \resizebox{\textwidth}{!}{%
  \begin{tabular}{M{3cm}|M{5cm}|M{3cm}|M{5cm}}
    \toprule
    \multicolumn{1}{c|}{Category}  & \multicolumn{1}{c|}{Category Description} & \multicolumn{1}{c|}{Related Commands} &  \multicolumn{1}{c|}{Example Task Description}  \\
    \midrule
    File System Manipulation  
    & Evaluate the knowledge of basic file system manipulation operation such as creating, deleting, copying, moving, compressing and listing files and directories.
    & mkdir, touch, zip, tar, ls, rm
    & List all files larger than 1MB inside the '/var/log' directory and write the list to a file named 'large\_files.txt' in the home directory.\\
    
    \midrule
    System Setting 
    & Evaluate the knowledge of system setting such as disk partition, OS version, user management.
    & df, useradd, groupadd, uname, chmod, whoami, chown
    & A user needs permission to read a file in '/var/private/info.txt'. Grant read access to all users. \\

    \midrule
    Process Running 
    & Evaluate the knowledge of processes management
    & renice, gcc, g++, python
    & Change the priority of the process with PID stored in /tmp/pidfile to a nice value of 10. \\

    \bottomrule
  \end{tabular}%
  }
\end{table*}

\begin{figure*}[t]
    \centering
    \includegraphics[width=0.99\linewidth]{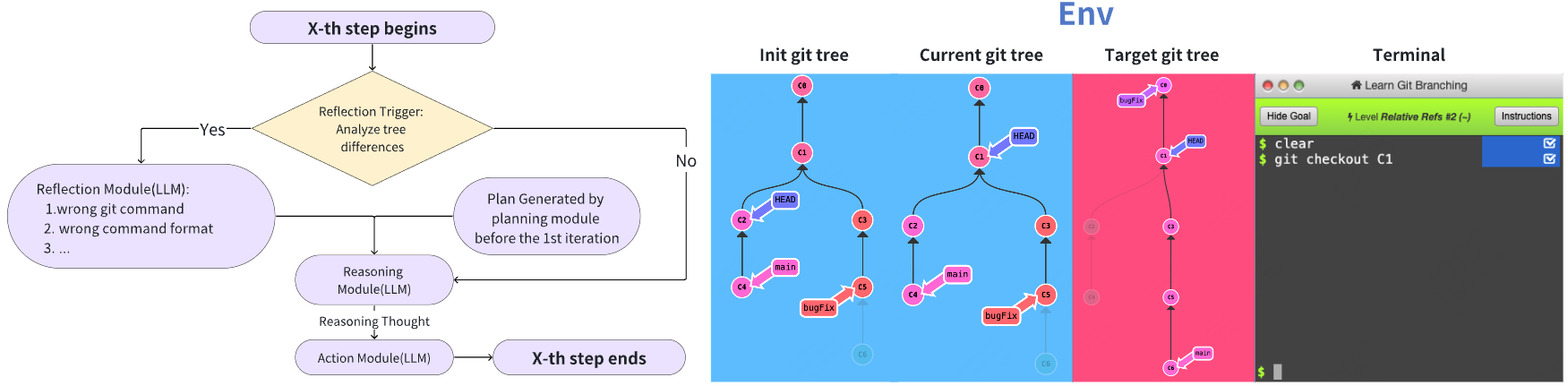}
    \caption{Illustration of OS-git task}
    \label{fig:coq_lean_isabelle}
    \vspace{-0.2cm}
\end{figure*}

\subsection*{Robot Cooperation}
The Robot Cooperation dataset evaluates agents' planning, reasoning, action, and reflection capabilities in multi-robot collaboration tasks. The dataset includes \textbf{100 tasks}, designed to benchmark performance in robot planning scenarios.

\textbf{Framework and Dataset Construction.}  
The dataset is built upon the RoCoBench environment framework, which provides an environment simulator and reward mechanisms for multi-robot collaboration tasks. We extended the original task set by introducing sequential constraints and leveraging random seed variations to generate diverse task instances.
\begin{itemize}
    \item \textbf{Task Extension:} Sequential constraints were added to existing tasks, making them more complex. Examples include:
    \begin{itemize}
        \item \textit{Sweep Floor Task:} Added order constraints. In the \textit{Sweep RGB} task, robots must first sweep the Red Cube into the dustpan and dump it into the bin, followed by the Green Cube, and finally the Blue Cube.
        \item \textit{Arrange Cabinet Task:} Introduced sequential object retrieval. In the \textit{CabinetCup} task, robots must first place the Cup on the Cup Coaster, followed by placing the Mug on the Mug Coaster.
        \item \textit{Sandwich Task:} Expanded with additional recipes requiring more planning steps.
    \end{itemize}
    \item \textbf{Task Instances:} Random seed variations in the RoCoBench environment were used to create different initial states, generating 100 unique task instances. Each instance was manually verified to ensure it has a correct solution, ensuring robustness and reliability for model evaluation.
\end{itemize}

\textbf{Reward Mechanism Improvements.}  
To better evaluate model capabilities, we proposed new reward methods tailored to the characteristics of the extended tasks:
\vspace{-0.2cm}
\begin{itemize}
    \item Tasks were divided into smaller sub-tasks with rewards granted for completing each sub-task in sequence.
    \item For example, in the \textit{Sweep RGB} task, rewards are distributed as $\frac{1}{3}$ for successfully completing each step (e.g., sweeping the Red Cube, Green Cube, and Blue Cube in order). This approach incentivizes correct sequencing and provides granular feedback on agent performance.
    \item These new reward methods ensure even smaller models can effectively receive feedback, improving evaluation sensitivity.
\end{itemize}

\textbf{Model Differentiation Enhancements.}  
To further enhance the differentiation capability of the models, we adopt a method where multiple actions are proposed within a single interaction. This approach, combined with a constraint on the number of timesteps, improves the differentiation among models. By allowing the agent to plan and propose multiple actions at once, we can better assess the agent's planning and reasoning abilities. The constraint on timesteps ensures that the agent must efficiently utilize its planning capabilities within a restricted timeframe, thereby providing a clearer distinction between the performance of different models.

\subsection*{Prompt Example}
\begin{lstlisting}[language=Python]
prompt_system_planning = """ 
Welcome to the Online Shopping Challenge! Four LLM agents are working together to do web-shopping tasks step by step (planning -> reasoning -> acting -> reflecting). They are responsible for planning, reasoning, acting, and reflecting respectively. 
You are the first llm agent, who is a helpful web-shopping guidance assistant in charge of planning. 
Your role is to assist players by generating strategic plans based on the game's instructions.

Here is how the game is structured:
- Each round, you will be given an instruction that describes the objective need to achieve.
- Based on the instruction, you are to generate a clear and brief strategic plan.
- Your plan will be used to guide other agents through the shopping site efficiently.
- If there is no response click[Buy Now] within 15 actions, the game fails.

Your Responsibilities:
- Analyze the original problem and break it into clear, actionable steps.
- Ensure the steps are logically ordered and comprehensive for achieving the goal.
- Use concise language, focusing only on the key actions needed to complete the task successfully.

OUTPUT FORMAT:
Keep your response concise and structure:
  Strategic Plan: (A list of sequential steps to achieve the objective)
	Step 1: ...
	Step 2: ...
	Step 3: ...
(Add more steps as necessary, but keep it streamlined and goal-oriented)

Enclose the plan with three backticks ```.

For example:
"""
\end{lstlisting}

\begin{lstlisting}[language=Python]
prompt_system_reasoning = """
Welcome to the Online Shopping Challenge!
Four llm agents are working together to do web-shopping tasks step by step(planning -> reasoning -> acting -> reflecting). They are responsible for planning, reasoning, acting and reflecting respectively.
You are the second LLM agent, who is a helpful web-shopping guidance assistant in charge of reasoning.
Your reasoning thought will guide the acting agent in making informed decisions. You should generate a thought that will be used as part of the PROMPT for acting agents.

In each round, following information will be given to you:
1. CURRENT OBSERVATION AND AVAILABLE ACTIONS
2. PLANNING STRATEGY
3. HISTORICAL ACTIONS
4. REFLECTION INFORMATION(if any)

Here is what you need to focus on:
- Every round, you will receive updated information about the shopping scenario, including the current observation, available actions, planning strategy, and past actions.
- Based on the current state, develop a clear thought process to guide the acting agent's next move.
- Ensure your response is directly actionable and aligns with the goal of achieving success in the game within 15 actions.
- If the game is nearing the interaction limit, prioritize quick decisions over perfect matches to ensure a [Buy Now] action happens promptly.
- When you determine that a sufficient match is found (even if not perfect), guide the acting agent to click [Buy Now] immediately.

OUTPUT FORMAT:
Based on the provided observation and available actions, generate a clear and brief thought in one sentence that outlines your analysis and considerations for the next move.
Note: Please surround the reasoning content you generated with three backticks. That is:
"""
\end{lstlisting}

\begin{lstlisting}[language=Python]
prompt_system_action = """
Welcome to the Online Shopping Challenge!
Four llm agents are working together to do web-shopping tasks step by step(planning -> reasoning -> acting -> reflecting). They are responsible for planning, reasoning, acting and reflecting respectively. 
You are the third LLM agent, who is a helpful web-shopping guidance assistant in charge of acting.
As an acting agent, your role is to integrate various elements such as the instruction, the current state, historical actions, strategic planning, and current reasoning to recommend the best possible action for the next step.

In each round, the following information will be given to you:
1. ORIGINAL PROBLEM
2. PLANNING STRATEGY
3. HISTORICAL ACTIONS
4. CURRENT REASONING

Your Role:
- Each round, you will receive updated information, including the current observation, available actions, strategic plan, reasoning, and past actions.
- Based on this information, decide and respond with the best possible action to move closer to completing the objective.
- Actions you can perform:
	Search if a search bar is available.
	Click one of the provided clickable buttons.
- Follow the reasoning closely, but only deviate if you are confident that your choice is better.

Important Rules:
- You must click [Buy Now] as soon as you are confident that a suitable match has been found to avoid exceeding the 15-round limit.
- If no valid action is available, perform no action and wait for the next round.
- Ensure the clicked value exactly matches the available options, including case sensitivity and punctuation.
- Attention: Although you need to click to buy as early as possible to get rewards, remember that you must click on a product before clicking to buy; 
			 if you click to buy without clicking on the product, you will receive 0 rewards.

OUTPUT FORMAT:
Use the following formats for your action:
	- searching: search [keywords]
	- clicking: click [value]
- For example: click [b06xdg8xfx]
- Keywords in search is up to you, but value in click must be a value in the list of available actions.
- The value must exactly match the original text, including case sensitivity (uppercase/lowercase) and all symbols/punctuation.

Note: Please surround the action content you generated with three backticks. That is:
"""
\end{lstlisting}

\begin{lstlisting}[language=Python]
prompt_system_reflection = """
Welcome to the Online Shopping Challenge!
Four llm agents are working together to do web-shopping tasks step by step(planning -> reasoning -> acting -> reflecting). They are responsible for planning, reasoning, acting and reflecting respectively. 
You are the fourth llm agent in charge of reflecting. Your role is to reflect on whether there was an error in the previous reasoning and action sequence.
Remember, your clear and brief reflection will be used as part of the PROMPT for the later agents to guide them to make wise decisions and succeed in the game.

In each round, the following information will be given to you:
1. ORIGINAL PROBLEM
2. HISTORICAL REASONINGS
3. HISTORICAL ACTIONS

Here is your role:
As an LLM Agent, your role is to reflect on the recent outcomes and consider the following points:
1. Identify why the current result is unsatisfactory. Explore factors such as inadequate search queries, irrelevant clicks, or repeated useless actions.
2. Evaluate the effectiveness of past actions and thoughts. Were there missed signals or incorrect assumptions?
3. Propose improvements for the next steps. Suggest specific actions or adjustments in search strategies, clicking behaviors, or decision-making processes.
4. Consider the overall goal of achieving successful purchases within the game's constraints. How can future actions better align with this objective?
Use these as a guide, and generate a plan for the next reasoning and action steps. Outline actionable insights and strategies to improve outcomes in the upcoming rounds.

OUTPUT FORMAT:
- You should carefully examine reasoning history and action history to find out where things may have gone wrong, summarize where they went wrong.
- Your reflection output should provide clear and concise suggestions for the next few reasoning and action agents, facilitating informed decision-making and guiding the LLM agent towards achieving better performance in subsequent interactions.
- Ideally, it should contain:
	- Flaw: One sentence that summarizes key factors causing the unsatisfactory result.
	- Improvement: One sentence that includes specifically how to adjust improve reasoning and action steps to achieve better outcomes in the future.

Note: Please enclose the flaw and improvement with three backticks:
"""
\end{lstlisting}

\end{document}

%% file: Table/mainResult.tex
\begin{table*}[t]
    \caption{Results Across Datasets. Metrics for baseline models are highlighted in blue. The evaluation covers 9 models across 7 tasks. Results marked with `*` below each dataset indicate the best-performing combinations computed based on Shapley Value.}
  \label{tab:consolidated_results_vertical}
  \vspace{-0.6cm}
  \centering
  \renewcommand{\arraystretch}{0.85}
  \begin{flushleft}
  \end{flushleft}
  \definecolor{headergray}{RGB}{240,240,240}
  \definecolor{highlightblue}{RGB}{230,240,255}
  \definecolor{highlightgray}{RGB}{248,248,248}
    {
    \scriptsize
    \begin{tabularx}{\textwidth}{>{\centering\arraybackslash}p{1.5cm}%
                              >{\centering\arraybackslash}p{1.3cm}%
                              *{10}{>{\centering\arraybackslash}X}}
    \toprule
    \rowcolor{headergray}
    \textbf{Dataset} & \textbf{Metric} & \textbf{\texttt{\shortstack{Llama3\\8B}}} & \textbf{\texttt{\shortstack{Claude\\3.5}}} & \textbf{\texttt{\shortstack{gpt-4o\\mini}}} & \textbf{\texttt{\shortstack{glm-4\\air}}} & \textbf{\texttt{\shortstack{qwen2.5\\32B}}} & \textbf{\texttt{\shortstack{Mistral\\8X7B}}} & \textbf{\texttt{\shortstack{Mistral\\7B}}} & \textbf{\texttt{\shortstack{gpt-4\\turbo}}} & \textbf{\texttt{\shortstack{doubao\\pro-4k}}} & \textbf{\texttt{\shortstack{Llama3\\70B}}} \\
    \midrule

    \multirow{6}{*}{\textbf{\shortstack{Online\\Shopping\\ \textit{Acc: 43.31*}}}} 
    & P            & \cellcolor{highlightblue}--      & -0.004   & 0.071 & \textbf{0.106} & -0.030  & -0.048  & 0.024  & 0.026   & \underline{0.071}   & -0.028  \\
    & R            & \cellcolor{highlightblue}--      & 0.019   & -0.025 & \textbf{0.077} & 0.004  & \underline{0.036} & 0.016  & -0.074  & 0.011   & 0.005  \\
    & A            & \cellcolor{highlightblue}--      & 0.056   & 0.068  & -0.059 & \textbf{0.156} & 0.080  & 0.004  & 0.014   & -0.045  & \underline{0.117}  \\
    & F            & \cellcolor{highlightblue}--      & -0.009   & \underline{-0.003} & -0.011 & -0.021  & -0.015 & -0.022  & \textbf{0.024}  & -0.040  & -0.030  \\
    & \cellcolor{highlightgray} Acc (\%)      & \cellcolor{highlightblue}26.27   & \cellcolor{highlightgray} 32.43   & \cellcolor{highlightgray} \underline{37.43}   & \cellcolor{highlightgray} \textbf{37.50}  & \cellcolor{highlightgray} 37.18   & \cellcolor{highlightgray}31.67   & \cellcolor{highlightgray}28.48    & \cellcolor{highlightgray}25.31  & \cellcolor{highlightgray}25.95   & \cellcolor{highlightgray}32.61  \\
    \midrule

    \multirow{6}{*}{\textbf{\shortstack{Navigation\\Planning\\ \textit{Acc: 74.42*}}}} 
    & P            & \cellcolor{highlightblue}--      &  0.000   & 0.006 & 0.001 & -0.002  & \underline{0.021} & \textbf{0.023}  & 0.008  & 0.001   & -0.009  \\
    & R            & \cellcolor{highlightblue}--      & \underline{0.030}   & 0.027 & -0.008 & 0.012  & -0.035  & \textbf{0.055}  & 0.014  & -0.003  & -0.019  \\
    & A            & \cellcolor{highlightblue}--      & \textbf{0.106}   & 0.081 & 0.005 & \underline{0.099}  & 0.048 & 0.042  & \underline{0.099}  & -0.051  & 0.046  \\
    & F            & \cellcolor{highlightblue}--      &  -0.006  & 0.002  & -0.021 & \textbf{0.018}  & -0.029 & \underline{0.007}  & 0.004 & -0.033  & -0.011  \\
    & \cellcolor{highlightgray} Acc (\%)      & \cellcolor{highlightblue}58.70   & \cellcolor{highlightgray} \textbf{71.90}   & \cellcolor{highlightgray} 70.29   & \cellcolor{highlightgray} 61.91  & \cellcolor{highlightgray} 68.26   & \cellcolor{highlightgray} 64.45   & \cellcolor{highlightgray} \underline{71.48}    & \cellcolor{highlightgray}71.23  & \cellcolor{highlightgray}50.90  & \cellcolor{highlightgray}59.32  \\
    \midrule

    \multirow{6}{*}{\textbf{\shortstack{Ticket\\Ordering\\ \textit{Acc: 67.18*}}}}
    & P            & \cellcolor{highlightblue}--      & 0.003   & 0.032 & -0.195 & 0.119  & \textbf{0.183}  & -0.111  & -0.043  & \underline{0.151}  & 0.004  \\
    & R            & \cellcolor{highlightblue}--      & 0.186   & \underline{0.243} & 0.172  & 0.181  & 0.054  & -0.070  & \textbf{0.301}  & -0.001  & 0.089  \\
    & A            & \cellcolor{highlightblue}--      & \textbf{0.217}   & \underline{0.049} & -0.020  & -0.000 & -0.083 & -0.020  & 0.028 & 0.006  & -0.275  \\
    & F            & \cellcolor{highlightblue}--      & 0.024   & 0.005  & -0.006 & \underline{0.043}  & -0.011 & 0.002   & \textbf{0.058}  & -0.027  & -0.001  \\
    & \cellcolor{highlightgray} Acc (\%)      & \cellcolor{highlightblue}19.94   & \cellcolor{highlightgray} \textbf{62.85}   & \cellcolor{highlightgray} 51.82   & \cellcolor{highlightgray} 15.01   & \cellcolor{highlightgray} 54.25   & \cellcolor{highlightgray} 34.24   & \cellcolor{highlightgray} 0.00   & \cellcolor{highlightgray} \underline{54.37}   & \cellcolor{highlightgray} 32.88  & \cellcolor{highlightgray} 1.59  \\

    \midrule

    \multirow{6}{*}{\textbf{\shortstack{Math \\ \textit{Acc: 83.80*}}}}
    & P            & \cellcolor{highlightblue}/      & 0.038& \underline{0.067}& 0.056& 0.065& 0.005& -0.060& 0.048& \textbf{0.115}& 0.028 \\
    & R            & \cellcolor{highlightblue}/      & \textbf{0.131}& 0.021& 0.044& \underline{0.107}& 0.003& -0.000& 0.065& 0.059& 0.031\\
    & A            & \cellcolor{highlightblue}/      & 0.442& 0.343& 0.348& \underline{0.483}& 0.164& -0.044& \textbf{0.492}& 0.182& 0.327 \\
    & F            & \cellcolor{highlightblue}/      & \underline{0.042}& \textbf{0.043}& 0.005& 0.031& -0.014& -0.003& 0.022& -0.002& 0.006\\
    & \cellcolor{highlightgray}Acc (\%)      & \cellcolor{highlightblue}18.00    & \cellcolor{highlightgray}\underline{83.40}& \cellcolor{highlightgray}65.40& \cellcolor{highlightgray}63.20& \cellcolor{highlightgray}\textbf{86.60}& \cellcolor{highlightgray}33.80& \cellcolor{highlightgray}7.20& \cellcolor{highlightgray}80.60& \cellcolor{highlightgray}53.40& \cellcolor{highlightgray}57.20\\
    \midrule
    
    \multirow{6}{*}{\textbf{\shortstack{ATP\\ \textit{Acc: 86.79*}}}}
    & P            & \cellcolor{highlightblue}/ & 0.012& 0.018& 0.002& 0.018& \textbf{0.025}& 0.008& 0.012& 0.016& \underline{0.019}        \\
    & R            & \cellcolor{highlightblue}/  & \textbf{0.057}& -0.016& 0.005& \underline{0.030}& 0.018& 0.010& 0.027& 0.019& -0.056       \\
    & A            & \cellcolor{highlightblue}/  & \textbf{0.660}& 0.345& 0.161& 0.511& 0.039& -0.009& \underline{0.541}& 0.084& 0.125       \\
    & F            & \cellcolor{highlightblue}/   & \textbf{0.069}& 0.015& 0.021& \underline{0.037}& -0.011& -0.000& 0.023& 0.004& 0.011      \\
    & \cellcolor{highlightgray}Acc (\%)      & \cellcolor{highlightblue}5.45  & \cellcolor{highlightgray}\textbf{85.29}& \cellcolor{highlightgray}41.74& \cellcolor{highlightgray}24.32& \cellcolor{highlightgray}65.17& \cellcolor{highlightgray}12.61& \cellcolor{highlightgray}6.31& \cellcolor{highlightgray}\underline{65.77}& \cellcolor{highlightgray}17.72& \cellcolor{highlightgray}15.32     \\
    \midrule

    \multirow{6}{*}{\textbf{\shortstack{Robot\\Cooperation\\ \textit{Rwd: 92.63*}}}}
    & P            & \cellcolor{highlightblue}--      & \textbf{0.114}   & 0.075  & -0.024  & 0.090  & -0.005  & -0.014  & \underline{0.107}   & 0.021   & 0.043   \\
    & R            & \cellcolor{highlightblue}--      & \textbf{0.388}   & 0.189  & 0.116   & 0.268  & 0.033   & -0.000  & \underline{0.329}   & -0.004   & 0.152   \\
    & A            & \cellcolor{highlightblue}--      & \textbf{0.319}   & 0.196  & 0.008   & 0.277  & 0.052   & -0.021  & \underline{0.316}   & 0.204   & 0.175   \\
    & F            & \cellcolor{highlightblue}--      & \textbf{0.017}   & -0.003 & -0.012  & 0.003  & \underline{0.004}  & -0.001  & 0.001   & -0.012  & -0.008  \\
    & \cellcolor{highlightgray}Reward (\%)   & \cellcolor{highlightblue}8.85   & \cellcolor{highlightgray}\textbf{92.63}    & \cellcolor{highlightgray}54.43   & \cellcolor{highlightgray}17.60   & \cellcolor{highlightgray}72.59   & \cellcolor{highlightgray}17.27    & \cellcolor{highlightgray}5.17    & \cellcolor{highlightgray}\underline{84.18}   & \cellcolor{highlightgray}29.75    & \cellcolor{highlightgray}45.06    \\
    \midrule
    \multirow{6}{*}{\textbf{\shortstack{Operating\\System\\ \textit{Acc: 60.78*}}}} 
    & P            & \cellcolor{highlightblue}--      & \textbf{0.078}   & 0.042  & 0.047  & 0.060  & 0.032   & 0.004  & 0.050   & 0.065   & \underline{0.077}   \\
    & R            & \cellcolor{highlightblue}--      & \textbf{0.458}   & 0.305  & 0.305  & 0.311  & 0.194   & 0.047  & \underline{0.395}   & 0.215   & 0.313   \\
    & A            & \cellcolor{highlightblue}--      & \textbf{0.071}   & 0.065  & 0.041  & 0.053  & 0.009   & 0.019  & \underline{0.070}   & 0.060   & 0.040   \\
    & F            & \cellcolor{highlightblue}--      & -0.008   & \underline{0.020}  & 0.004  & \textbf{0.037}  & 0.001   & 0.019  & 0.005   & -0.006  & 0.012   \\
    & \cellcolor{highlightgray}Acc (\%)   & \cellcolor{highlightblue}0.98   & \cellcolor{highlightgray}\textbf{60.78}    & \cellcolor{highlightgray}44.12   & \cellcolor{highlightgray}40.71   & \cellcolor{highlightgray}47.06   & \cellcolor{highlightgray}24.51    & \cellcolor{highlightgray}9.80    & \cellcolor{highlightgray}\underline{52.94}   & \cellcolor{highlightgray}34.31    & \cellcolor{highlightgray}45.10    \\
    \midrule
    
  \end{tabularx}}
  \vspace{-0.4cm}
\end{table*}

%% file: Figure/radarChart/radar.tex
\begin{figure*}[t]
    \centering
    \includegraphics[width=0.95\linewidth]{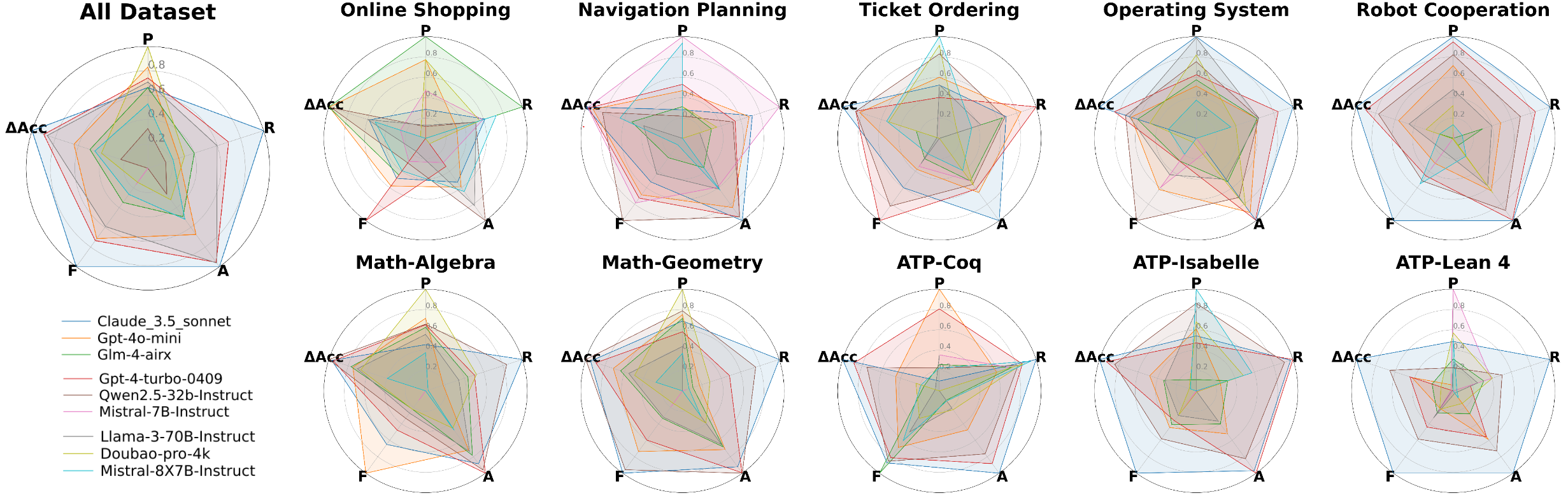}
    \vspace{-0.3cm}
    \caption{Radar plot comparing model performance across tasks with key contributions.}
    \label{fig:radar chart}
    \vspace{-0.2cm}
\end{figure*}
